\documentclass[preprint]{cas-sc}
\usepackage{amssymb}
\usepackage{amsmath}
\usepackage[switch]{lineno}
\usepackage{graphicx}
\usepackage{subfig} 
\usepackage{tabu}  
\usepackage{hyperref}
\usepackage{multirow}
\usepackage{dsfont}
\usepackage{stfloats}
\usepackage{cite}
\usepackage{booktabs}
\usepackage{algorithm}
\usepackage{algorithmic}
\usepackage{amssymb}
\usepackage{makecell}
\usepackage{caption}
\usepackage{orcidlink}
\usepackage{natbib}

\usepackage{hyperref}
\hypersetup{colorlinks=true}

\usepackage{xcolor}
\definecolor{wdcolor}{RGB}{0, 0, 255}

\definecolor{szcolor}{RGB}{154, 0, 0}

\def\tsc#1{\csdef{#1}{\textsc{\lowercase{#1}}\xspace}}
\tsc{WGM}
\tsc{QE}
\tsc{EP}
\tsc{PMS}
\tsc{BEC}
\tsc{DE}
\begin{document}
\let\WriteBookmarks\relax
\def\floatpagepagefraction{1}
\def\textpagefraction{.001}
\shorttitle{}
\shortauthors{Liu et~al.}

% \linenumbers
%\title [mode = title]{This is a specimen $a_b$ title}                      
% \tnotemark[1,2]

% \tnotetext[1]{This document is the results of the research
%    project funded by the National Science Foundation.}

% \tnotetext[2]{The second title footnote which is a longer text matter
%    to fill through the whole text width and overflow into
%    another line in the footnotes area of the first page.}

%\begin{frontmatter}



\title[mode = title]{MT-CYP-Net: Multi-Task Network for Pixel-Level Crop Yield Prediction Under Very Few Samples}

\author[1]{Shenzhou Liu}{}
\cormark[1]

\author[2]{Di Wang}
\cormark[1]

\author[3]{Haonan Guo}

\author[3]{Chengxi Han}

\author[1,4]{Wenzhi Zeng\orcidlink{0000-0003-0667-3604}}
\cormark[2]

\affiliation[1]{organization={State Key Laboratory of Water Resources Engineering and Management, Wuhan University},
            city={Wuhan},
            postcode={430072}, 
            state={Hubei},
            country={China}}

\affiliation[2]{organization={School of Computer Science, Wuhan University},
            city={Wuhan},
            postcode={430072}, 
            state={Hubei},
            country={China}}

\affiliation[3]{organization={State Key Laboratory of Information Engineering in Surveying, Mapping and Remote Sensing, Wuhan University},
            city={Wuhan},
            postcode={430079}, 
            state={Hubei},
            country={China}}

\affiliation[4]{organization={College of Agricultural Science and Engineering, Hohai University},
            city={Nanjing},
            postcode={211100}, 
            state={JiangSu},
            country={China}}   

\cortext[cor1]{Equal contribution}
\cortext[cor2]{Corresponding author}

\begin{abstract}
Accurate and fine-grained crop yield prediction plays a crucial role in advancing global agriculture. However, the accuracy of pixel-level yield estimation based on satellite remote sensing data has been constrained by the scarcity of ground truth data. To address this challenge, we propose a novel approach called the Multi-Task Crop Yield Prediction Network (MT-CYP-Net). This framework introduces an effective multi-task feature-sharing strategy, where features extracted from a shared backbone network are simultaneously utilized by both crop yield prediction decoders and crop classification decoders with the ability to fuse information between them. This design allows MT-CYP-Net to be trained with extremely sparse crop yield point labels and crop type labels, while still generating detailed pixel-level crop yield maps.
Concretely, we collected 1,859 yield point labels along with corresponding crop type labels and satellite images from eight farms in Heilongjiang Province, China, in 2023, covering soybean, maize, and rice crops, and constructed a sparse crop yield label dataset. MT-CYP-Net is compared with three classical machine learning and deep learning benchmark methods in this dataset. Experimental results not only indicate the superiority of MT-CYP-Net compared to previous methods on multiple types of crops but also demonstrate the potential of deep networks on precise pixel-level crop yield prediction, especially with limited data labels.

\end{abstract}

% \begin{highlights}
% \end{highlights}


%\end{frontmatter}


\maketitle

\section{Introduction}
\label{sec:Intro}

%%       \citet{<label>} ==> Jones et al. (2015)
%%       \citep{<label>} ==> (Jones et al., 2015)

Accurate large-scale high-resolution crop yield prediction is the core task for precise agriculture, for its  significant influence on food security, economy, and agricultural development. With accurate predictions, government and international organizations can make effective agricultural policies; agricultural insurance companies can design accurate and fair agricultural insurance products; farmers can make informed management decisions \citep{benamiuniting2021}. Crop yield is determined by crop genotype as well as various environmental conditions \citep{eltaher_gwas_2021}. However, precisely modeling these intricate physiological processes and monitoring diverse environmental factors and crop conditions across large areas are both challenging tasks. These complexities present significant obstacles to achieving accurate, high-resolution, and large-scale crop yield predictions.

The approaches for crop yield prediction can be classified into three main categories: mechanistic crop growth models, semi-empirical light energy utilization models, and data-driven models \citep{debaeke2023126677}. Mechanistic crop growth models estimate crop yield by modeling the development of crops and their interactions with the environment and management practices \citep{RWilliams1989, DEWIT2008414}. However, they require substantial on-site data for model calibration and insufficiently consider extreme environmental factors such as floods and lodging 
\citep{LUO2023103711}. On the other hand, semi-empirical light energy utilization models focus on estimating the total primary productivity (GPP) of crops based on the photosynthesis model and converting aboveground biomass into crop yield using the Harvest Index (HI) \citep{YU2024114301}. Although these models require fewer parameters, they overlook the comprehensive impacts of management and environmental conditions and provide relatively low accuracy \citep{YUAN2016702}. In summary, mechanistic crop growth models and semi-empirical models emphasize modeling the dynamics of crop development, lacking comprehensive consideration for the complex impacts of environmental factors. Therefore, they have built-in limitations in large-scale and high-resolution crop yield prediction.

\begin{figure*}[t]
    \centering
    \includegraphics[width=0.8\linewidth]{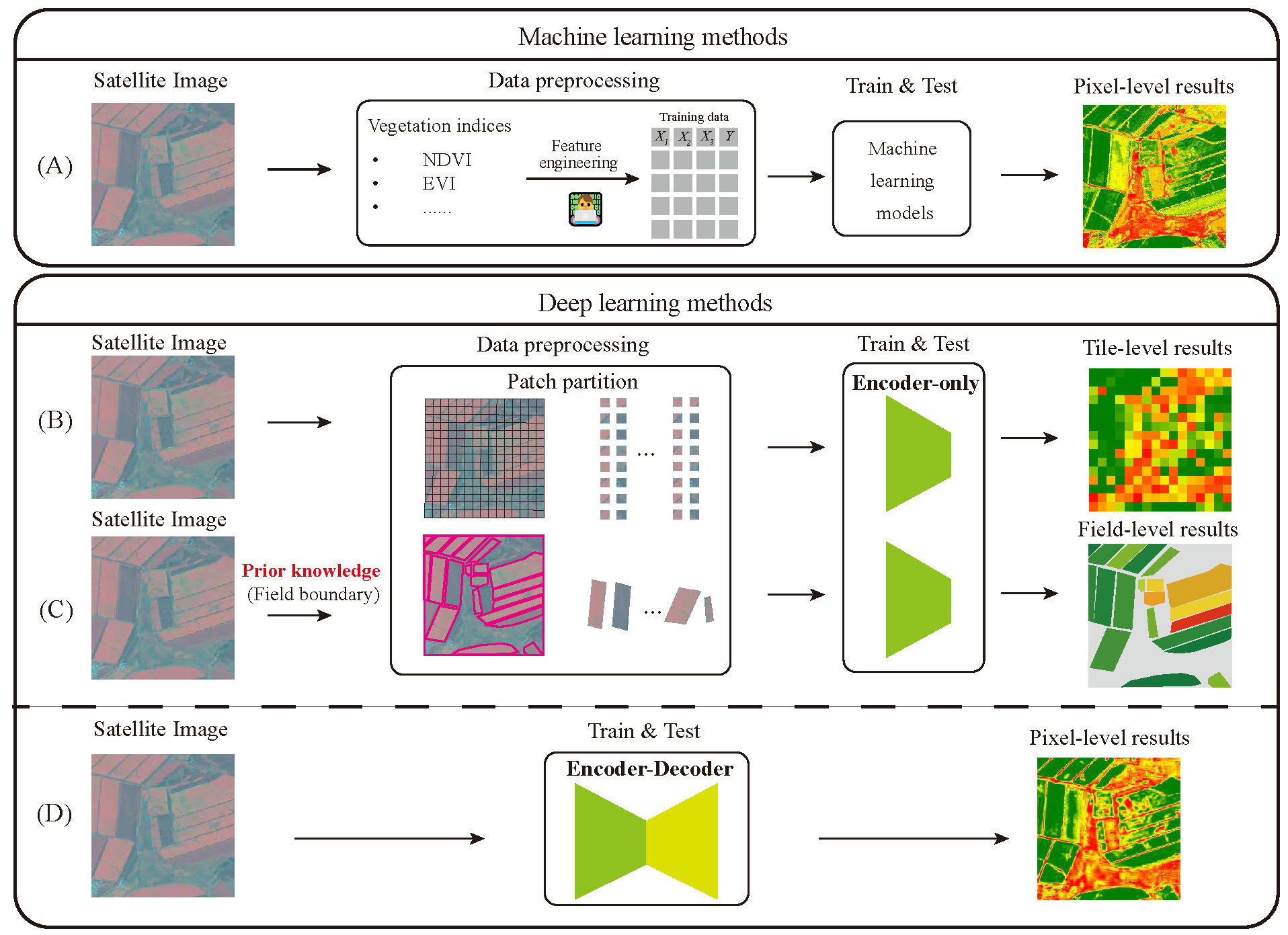}
    \caption{Different data-driven methods for crop yield prediction. (A) pixel-level machine learning methods. (B) tile-level deep learning methods. (C) field-level deep learning methods. (D) pixel-level deep learning methods.}
    \label{fig:intro}
\end{figure*}

In contrast, data-driven models simply focus on capturing the connections between high-dimension data and crop yield, providing flexible and efficient ways for crop yield prediction. These models can use diverse input data to capture the impacts of various factors and can be mainly categorized into machine learning methods and deep learning methods \citep{Alexandros}. Machine learning methods entail manually designed pixel-level features, including vegetation indices, which are subsequently inputted into machine learning models to derive pixel-level yield prediction results\citep{CLARKE2024108716,deFreitas2024} (Fig. \ref{fig:intro}A). Deep learning methods automatically capture deep features from input data and predict the crop yield through well-designed neural networks, and the most commonly used method is CNN \citep{Alexandros}. Given CNN's robust capacity to extract spatial semantic information, recent studies have demonstrated CNN outperforms classical machine learning methods in tile-level and field-level crop yield prediction using satellite images \citep{SAGAN2021265, YANG2019142}.

The tile-level process divides remote sensing images into smaller patches and conducts patch-wise regression for yield prediction (Fig. \ref{fig:intro}B), while the field-level process entails segmenting images according to field boundaries and then predicting the crop yield for each field  (Fig. \ref{fig:intro}C). However, according to our literature survey, CNN’s potential in pixel-level crop yield prediction has not been fully explored. \citet{9856943} firstly cast the problem of yield prediction as a dense prediction problem and highlighted the superiority of encoder-decoder deep learning models over traditional machine learning models (Fig. \ref{fig:intro}D). However, their study relies on dense high-resolution crop yield maps from high-precision harvesters. Due to the high annotation cost, their method is difficult to validate across a wider range of crop types and larger geographical areas for broad application and scalability. 

Therefore, to reduce data collection costs, this study would like to explore achieving precise pixel-level dense prediction on large-scale regions with fewer crop yield annotation samples. However, it is expected that the model performance is inevitably degraded when training with very few labels. One promising direction to address these challenges is multi-task learning (MTL), an approach to improve generalization ability by simultaneously leveraging the knowledge from different tasks \citep{Caruana1997}, which is particularly effective in data-limited conditions \citep{Moscato2023}. MTL has been successfully applied to simultaneously predict multiple crop physiological parameters to improve crop yield prediction accuracy, such as crop yield and protein content \citep{Zhuangzhuang}, along with crop yield and harvest level \citep{agriculture14040513}. For pixel-level yield prediction, the combination of semantic segmentation and regression tasks has been proven to be more effective \citep{8578175}. Therefore, in our consideration, integrating crop classification tasks into crop yield prediction models may present an opportunity for better crop yield prediction in data-limited conditions.

In this work, to achieve the MTL of dense crop yield prediction and classification, we simultaneously collect crop yield and category data on different outdoor locations. Then, we develop a multi-task encoder-decoder CNN model, called MT-CYP-Net (Multiple Task Crop Yield Prediction Network). This structure enables the network training can be improved by simultaneously encompassing the benefits from various task modeling, improving the performance on both dense crop yield prediction and pixel-level recognition tasks, even if with few samples.

The main contributions of this paper can be summarized as: 

(1) We develop the first end-to-end multi-task framework, where the classification and yield prediction tasks are jointly optimized for efficiently achieving pixel-level crop yield prediction on large-scale regions under very few crop yield data labels. 

(2) We collect multiple task data, including satellite images as well as both crop yield and crop type labels in real field conditions, where the crop yields are annotated at point-level on a small number of positions to reduce data acquisition costs. 

(3) We conduct quantitative and qualitative experiments by adopting diverse data types and various spectral band combinations. The results indicate that our model achieves excellent precision-efficiency trade-off on multiple crop types compared with classic machine learning methods and existing advanced deep learning-based networks, and enables efficient crop yield mapping on large-scale scenarios.

\section{Study area}
\begin{figure*}[t]
    \centering
    \includegraphics[width=1\linewidth]{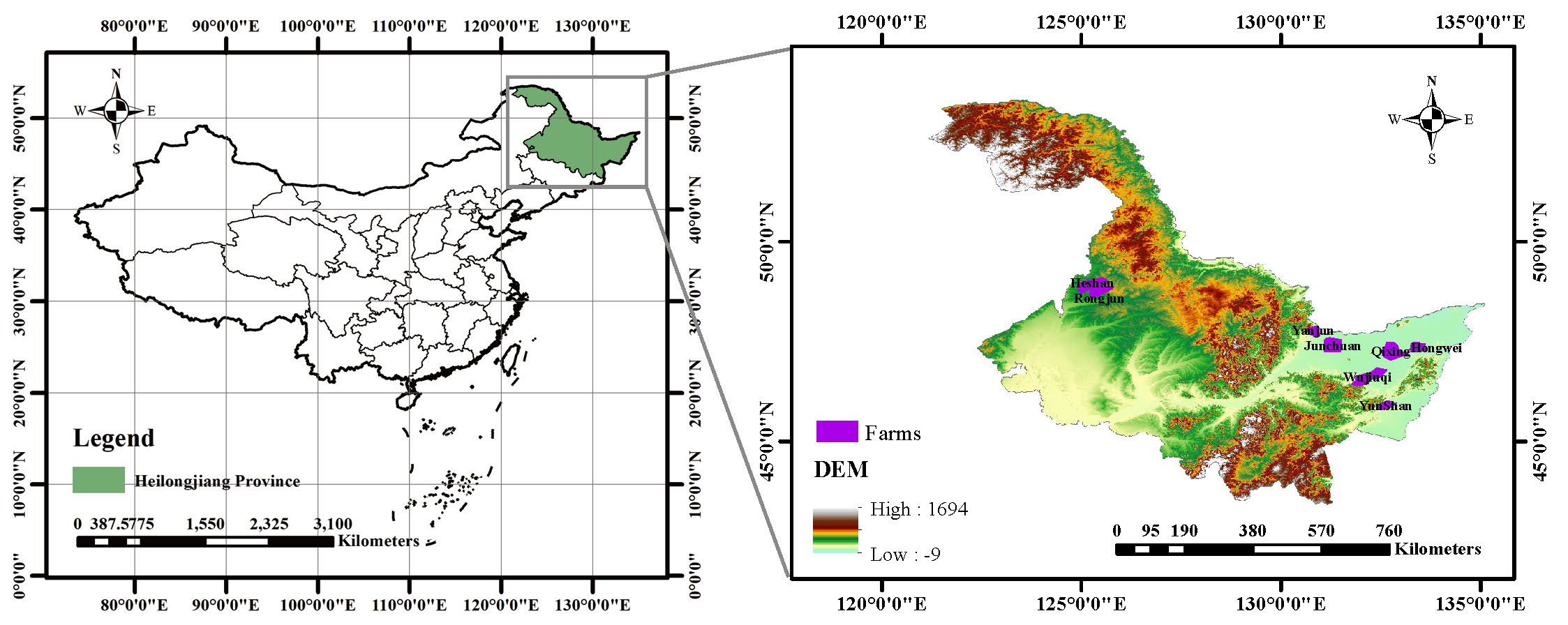}
    \caption{The study area of eight farms where crop yield and crop type data were collected.}
    \label{fig:map}
\end{figure*}

The data used in this study were collected from 8 farms in Heilongjiang Province, China, in 2023, namely Heshan, Hongwei, Junchuan, Qixing, Rongjun, Yanjun, Yunshan, and Wujiuqi (Fig. \ref{fig:map}). The area of these farms ranges from 300 to 1000 $\text{km}^2$ and mainly cultivate grain crops such as rice, maize and soybean. Maize and soybean are typically planted in April and harvested in October, while rice is typically transplanted in May and harvested in October.

\subsection{Crop yield and crop type data}

\begin{figure*}[t]
    \centering
    \includegraphics[width=1\linewidth]{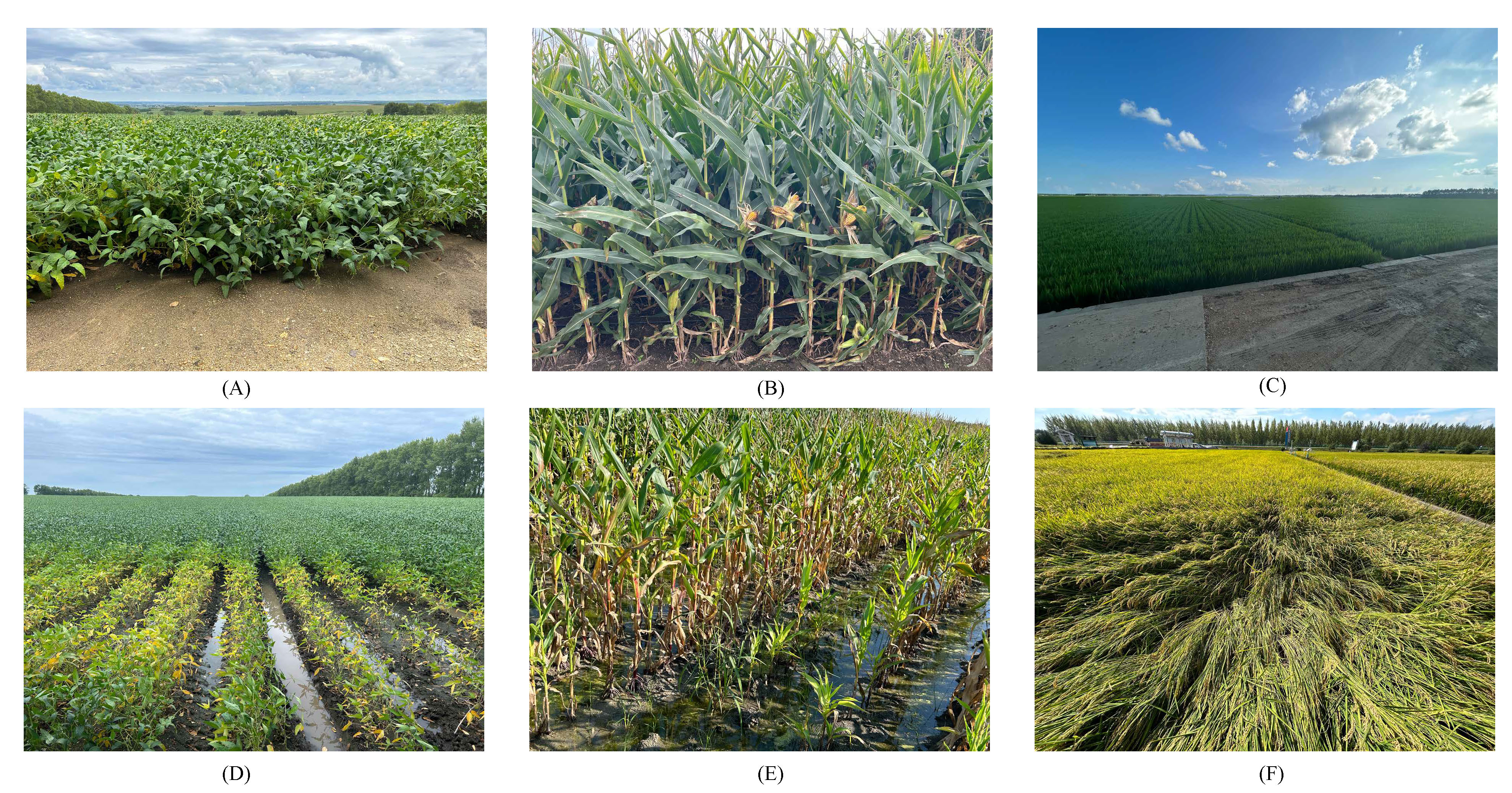}
    \caption{The representative crops in different growth conditions in late August. (A) Heathy soybean; (B) Heathy maize; (C) Heathy rice; (D) Unhealthy soybean; (E) Unhealthy maize; (F) Unhealthy rice.}
    \label{fig:crop_photo}
\end{figure*}

\textbf{Crop yield data.} The crop yield data was obtained in two ways: (1) We did an in-field survey in late August 2023, selecting crops with varying growth conditions (Fig. \ref{fig:crop_photo}), and sampled a plot around 1 $\text{m}^2$ to measure the crop yield per area. High-precision GPS technology was used to accurately record the location of each sampled plot, ensuring precise geospatial alignment between the field measurements and the satellite imagery. (2) Some crop yield labels were annotated by experienced farm technicians through manual visual interpretation on remote sensing images. Finally, the crop yields of different crop types were normalized to between 0-1 according to crop type for privacy reasons. 

\textbf{Crop type data.} The crop type data was provided by the local farms. In instances where some fields were labeled without a specific crop type, to best utilize the field annotations, we categorize them under the “other crop” class. For those objects that do not belong to crops such as water and roads, we categorize them into the “non-crop” class.  When an image is not fully labeled, we categorize the unlabeled area into the “unlabeled” class. Finally, the crop type annotations have 6 classes: rice, maize, soybean, other crop, non-crop, and unlabeled.

\subsection{Satellite data}

% \begin{table}[h]
% \caption{Details of different spectral bands for Sentinel-2 sensors. \citep{rs12142291}}
% \resizebox{\columnwidth}{!}{%
% \begin{tabular}{lllll}
% \hline
% \multirow{2}{*}{Name} & \multicolumn{2}{l}{Wavelength (nm)} & \multirow{2}{*}{Resolution (m)} & \multirow{2}{*}{Description} \\ \cline{2-3}
%  & Sentinel-2A & Sentinel-2B &  &  \\ \hline
% B01 & 442.7 & 442.3 & 60 & Coastal aerosol  \\
% B02 & 492.4 & 492.1 & 10 & Blue \\
% B03 & 559.8 & 559.0 & 10 & Green \\
% B04 & 664.6 & 665.0 & 10 & Red \\
% B05 & 704.1 & 703.8 & 20 & Vegetation red edge \\
% B06 & 740.5 & 739.1 & 20 & Vegetation red edge \\
% B07 & 782.8 & 779.7 & 20 & Vegetation red edge \\
% B08 & 832.8 & 833.0 & 10 & NIR \\
% B8A & 864.7 & 864.0 & 20 & Narrow NIR \\
% B09 & 945.1 & 943.2 & 60 & Water vapor \\
% B10 & 1373.5 & 1376.9 & 60 & SWIR - Cirrus \\
% B11 & 1613.7 & 1610.4 & 20 & SWIR \\
% B12 & 2202.4 & 2185.7 & 20 & SWIR \\ \hline
% \end{tabular}%
% }
% \label{tab:Sentinel-2-band}
% \end{table}

Sentinel-2 is a multispectral satellite tandem composed of two satellites, Sentinel-2A and Sentinel-2B, launched by the European Space Agency in 2015 and 2017. It can scan the Earth's surface with a revisit period of 5 days. The Sentinel-2 data products mainly include L1C and L2A levels, where L1C products have 13 bands and L2A products have 12 bands . L1C products have been processed by geometric and radiometric correction of reflectance data, while L2A products were further processed and mainly contain atmospheric corrected reflectance data. Some studies used L1C products for crop yield prediction \citep{ESTEVEZ2022112958, PERICH2023108824, Suarez2024}, while others also used L2A products \citep{DESLOIRES2023107807,rs11151745}. To test the robustness of models, we use both Sentinel-2 L1C and L2A products as the remote sensing image sources (https://registry.opendata.aws/sentinel-2/).

\subsection{Dataset description}

\begin{table}[t]
\caption{Sentinel-2 image dates, crop yield sample points number and types across eight farms in this study. $\textcolor[rgb]{0.976, 0.725, 0.443}{\bullet}$ soybean   $\textcolor[rgb]{0.133,0.710,0.443}{\bullet}$ rice $\textcolor[rgb]{0.188, 0.271, 0.188}{\bullet}$ maize    $\textcolor[rgb]{0.996, 0.925, 0.0}{\bullet}$ other crop}
\centering
\resizebox{0.6\columnwidth}{!}{%
\begin{tabular}{@{}lllll@{}}
\toprule
Farm & Date & Crop yield number & Image number & Crop types \\ \midrule
Junchuan & 2023/8/20 & 97 & 10 & $\textcolor[rgb]{0.133,0.710,0.443}{\bullet} \textcolor[rgb]{0.188, 0.271, 0.188}{\bullet} \textcolor[rgb]{0.976, 0.725, 0.443}{\bullet}$ \\
Qixing & 2023/8/20 & 223 & 17 & $\textcolor[rgb]{0.133,0.710,0.443}{\bullet} \textcolor[rgb]{0.976, 0.725, 0.443}{\bullet} \textcolor[rgb]{0.188, 0.271, 0.188}{\bullet}$ \\
Rongjun & 2023/8/19 & 97 & 37 & $\textcolor[rgb]{0.976, 0.725, 0.443}{\bullet} \textcolor[rgb]{0.188, 0.271, 0.188}{\bullet} \textcolor[rgb]{0.996, 0.925, 0.0}{\bullet}$ \\
Yanjun & 2023/8/20 & 169 & 16 & $\textcolor[rgb]{0.188, 0.271, 0.188}{\bullet} \textcolor[rgb]{0.976, 0.725, 0.443}{\bullet} \textcolor[rgb]{0.133,0.710,0.443}{\bullet} \textcolor[rgb]{ 0.996, 0.925, 0.0}{\bullet} $ \\
Yunshan & 2023/8/27 & 67 & 8 & $\textcolor[rgb]{0.133,0.710,0.443}{\bullet} \textcolor[rgb]{0.976, 0.725, 0.443}{\bullet} \textcolor[rgb]{ 0.188, 0.271, 0.188}{\bullet}$ \\
Heshan & 2023/8/19 & 557 & 79 & $\textcolor[rgb]{0.976, 0.725, 0.443}{\bullet} \textcolor[rgb]{0.188, 0.271, 0.188}{\bullet} \textcolor[rgb]{0.996, 0.925, 0.0}{\bullet}$ \\
Hongwei & 2023/8/27 & 11 & 6 & $\textcolor[rgb]{0.976, 0.725, 0.443}{\bullet} \textcolor[rgb]{0.188, 0.271, 0.188}{\bullet} $ \\
Wujiuqi & 2023/8/20 & 82 & 9 & $\textcolor[rgb]{0.976, 0.725, 0.443}{\bullet} $ \\ \bottomrule
\end{tabular}%
}
\label{tab:img_dates}
\end{table}

\begin{figure}[h]
    \centering
    \includegraphics[width=0.5\linewidth]{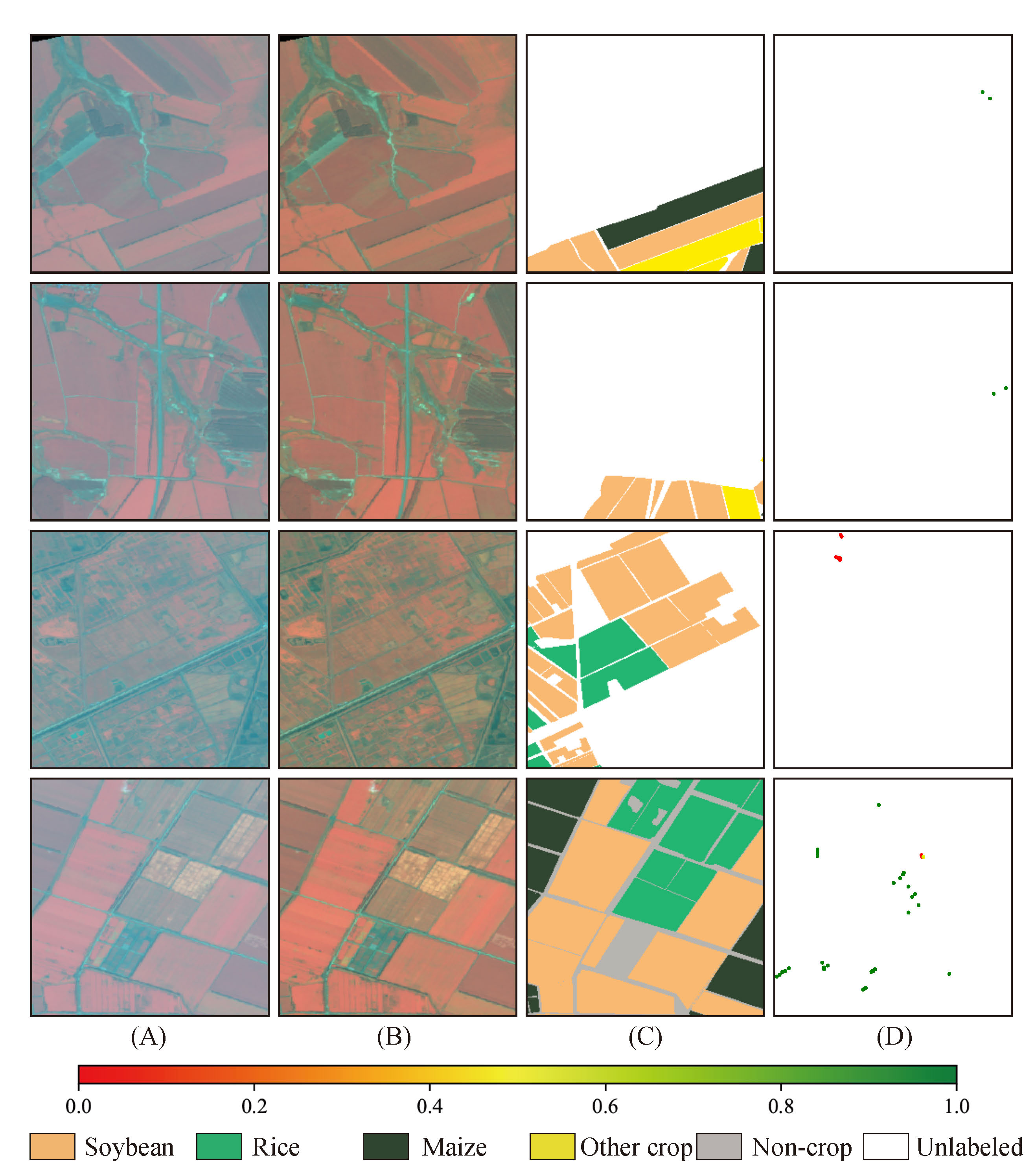}
    \caption{The visualization of our dataset. (A) Sentinel-2 L1C images displayed in pseudo color (NIR, Green, Blue); (B) Sentinel-2 L2A images displayed in pseudo color (NIR, Green, Blue); (C) Crop type labels; (D) Crop yield point labels.}
    \label{fig:dataset_visual}
\end{figure}

The dataset preprocessing involves several stages. First, we select the satellite images with less cloud cover, which were captured around August 2023 (Table. \ref{tab:img_dates}). Next, we employ cubic convolution interpolation to interpolate the 20 m and 60 m resolution bands to 10 m. Then, we rasterize the crop yield and crop type labels to match the resolution of Sentinel-2 imagery, ensuring alignment with the satellite data. Finally, the satellite images, crop yield raster and crop type raster were cropped into images with a width and height of 256 × 256 pixels using a sliding window approach with an overlap rate of 0.1.

We then aggregate all the images from the eight farms to create two datasets: L1C and L2A. Each dataset comprises 182 images and 1,859 crop yield points (as detailed in Table \ref{tab:img_dates}), where we only select the images with crop yield annotations. Next, we fix the random seeds and perform 10-fold cross-validation with a 9:1 split ratio, dividing the datasets into training and validation sets. It is important to note that the images in the L1C and L2A datasets have a one-to-one correspondence. Therefore, for each fold, the training and validation sets of both datasets (L1C and L2A) are in the same geographical locations (Fig. \ref{fig:dataset_visual}).

\section{Methodology}

\subsection{MT-CYP-Net}

\begin{figure*}[t]
    \centering
    \includegraphics[width=1\linewidth]{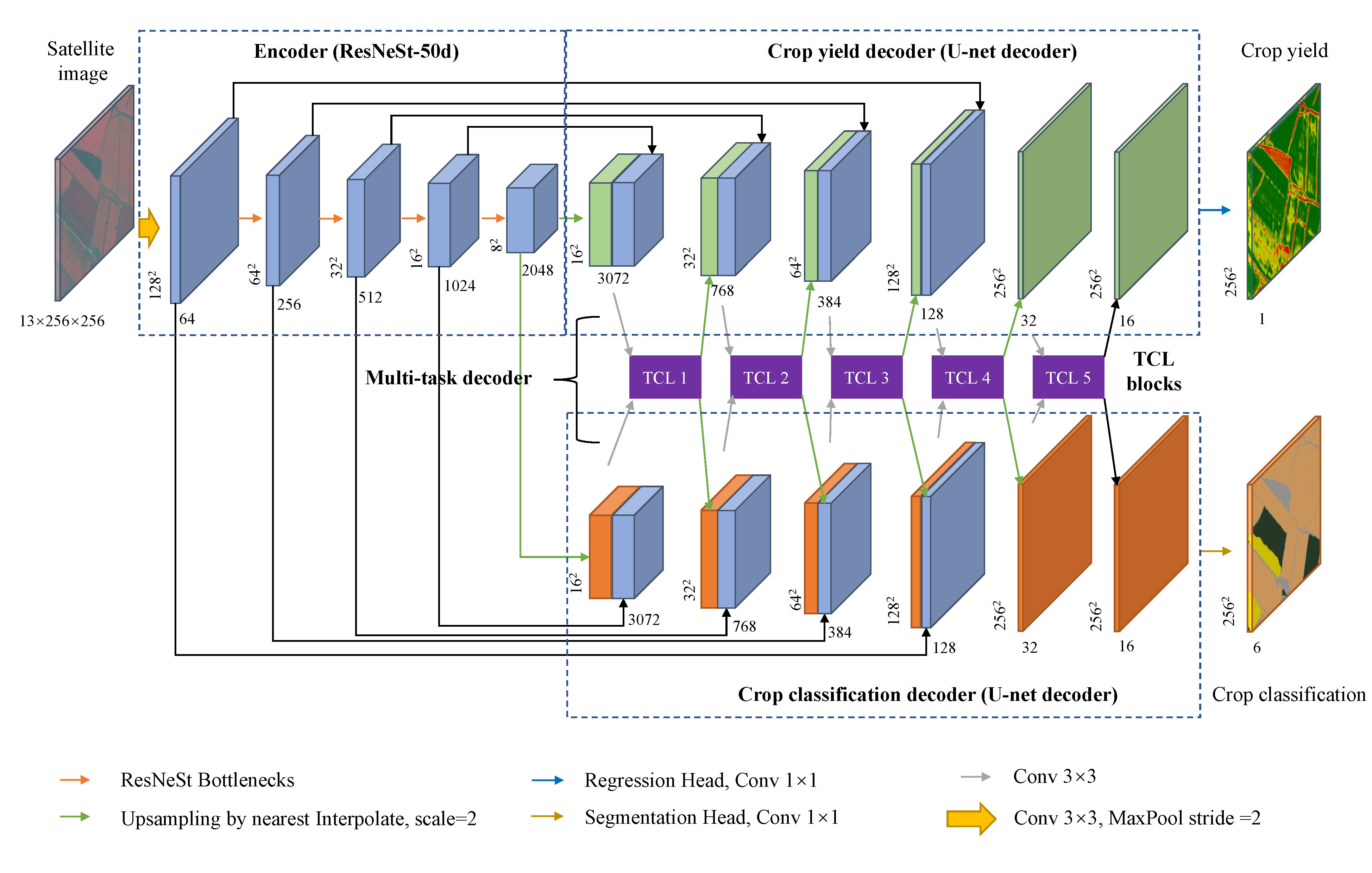}
    \caption{The overall architecture of the proposed MT-CYP-Net. The multi-task decoders comprise a crop yield decoder and a crop classification decoder.}
    \label{fig:network}
\end{figure*}

\begin{figure}[t]
    \centering
    \includegraphics[width=0.7\linewidth]{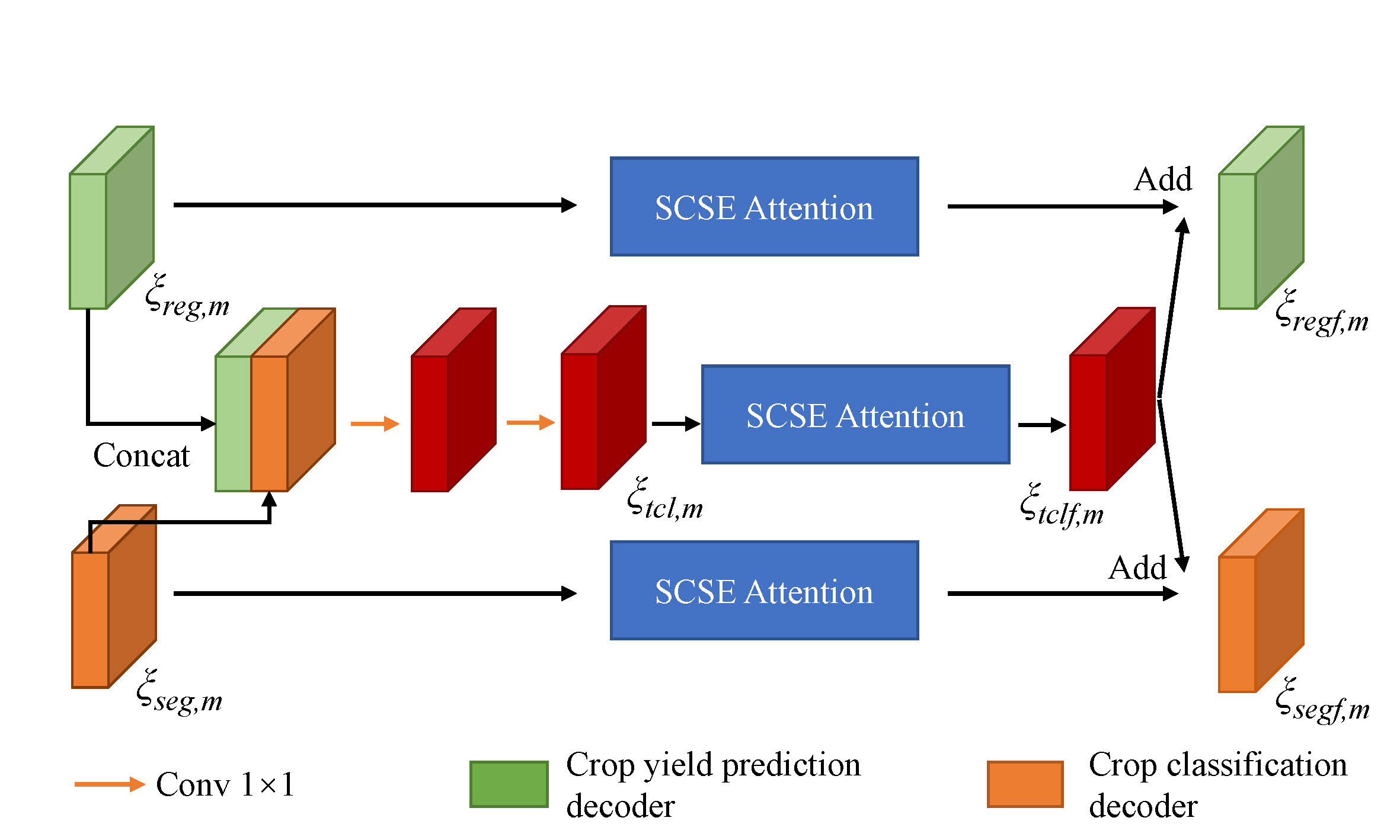}
    \caption{Detailed structures of the TCL block. }
    \label{fig:tcl}
\end{figure}

As mentioned earlier, pixel-level crop yield prediction can be regarded as a dense prediction problem. However, sparse labels often lead to unstable training of deep neural networks, while the multi-task learning paradigm helps mitigate this issue by enabling the model to learn more robust feature representations. Based on this insight, we propose the Multi-Task Crop Yield Prediction Network (MT-CYP-Net), which jointly predicts crop yield and crop type by single satellite imagery. The network architecture comprises three key components: (1) Image encoder, (2) Multi-task decoder, and (3) TCL blocks (see in Fig. \ref{fig:tcl}). 

Given its efficiency and suitability for small datasets \citep{rs13224533, Wang2023}, as well as has been proven effective in crop yield prediction in the previous study, the Unet \citep{unet} is employed as the encoder-decoder framework of the proposed MT-CYP-Net. This network utilizes a shared encoder and simultaneously generates both crop yield predictions and classification results by a regression decoder and a segmentation decoder. Additionally, to facilitate the interaction between crop yield prediction and crop classification tasks, we introduce the Task Consistency Learning (TCL) block \citep{9402788} at each upsampling layer of the network. The following text will provide a detailed introduction for each component.

\subsection{Image encoder}

Following the common practices of CNN models, we select two classical CNN architectures DenseNet-161 \citep{Huang_2017_CVPR} and ResNet-50 \citep{He_2016_CVPR}. In addition, the recently advanced ResNeSt-50d \citep{ResNeSt}, which uses the split attention module to improve diverse feature representations, is also considered.

\subsection{Multi-task decoder}
The crop yield prediction can be regarded as a dense regression task, and crop classification is a semantic segmentation task. Therefore, the multi-task decoder is composed of two parallel decoders: a crop yield decoder (for regression) and a crop classification decoder (for segmentation).

Both decoders follow the architecture of the Unet decoder \citep{unet}, which progressively restores the spatial resolution of the feature map downsampled by the image encoder, ultimately generating an output matching the input image size. Each decoder consists of upsampling and convolutional layers and performs feature fusion with corresponding encoder layers through skip connections. The only difference between the two decoders is the channel number of the segmentation head, where the crop yield decoder is 1 and the crop classification decoder is 6 (including unlabeled).

\subsection{TCL block}

\begin{figure*}[t]
    \centering
    \includegraphics[width=1\linewidth]{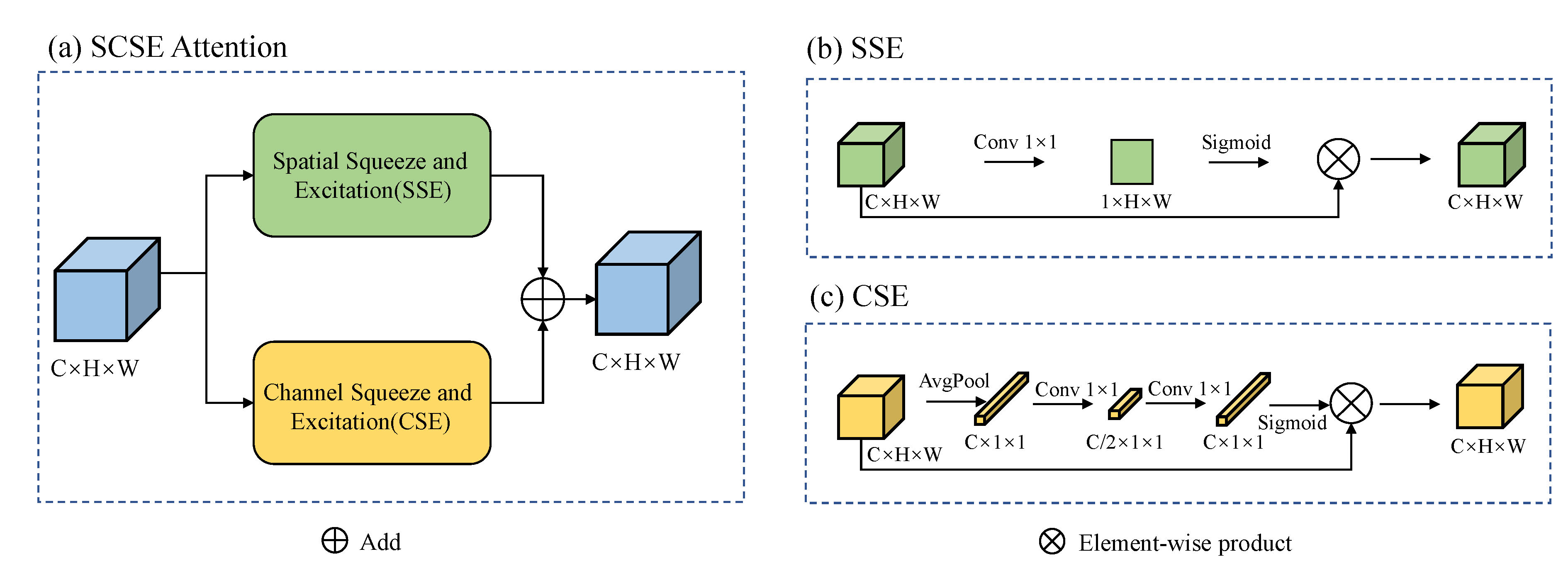}
    \caption{The structures of the (a) SCSE attention. It consists of two components: (b) Spatial Squeeze and Excitation (SSE), and (c) Channel Squeeze and Excitation (CSE). }
    \label{fig:SCSE}
\end{figure*}

In view of the scale invariance of crop category and yield, at each upsampling layer of the multi-task decoder, TCL blocks are employed to facilitate information fusion between the crop yield prediction decoder and crop classification decoder, strengthening crop perception under multi-scale features (see Fig. \ref{fig:network}). The TCL block primarily consists of two key components: SCSE (Spatial and Channel Squeeze and Excitation) attention \citep{SCSE} and feature sharing component. Since crops possess strong spatial contextual correlation, i.e., crops in the same field exhibit spatial similarity, and unique spectral properties, meaning the reflectances of crops are changed by wavelength, while different crops present various spectral profiles. The SCSE attention module is able to enhance the model's ability by refining spatial and channel-wise features (Fig. \ref{fig:SCSE}). In addition, the feature sharing component receives feature maps from two decoders and generates fusion feature maps, which allows the model to combine complementary task-specific information.

Specifically, in the \textit{m}-th TCL block, the feature maps $\xi _{reg,m}$ and $ \xi _{seg,m}$ from the regression and segmentation branches are first combined through channel concatenation. The concatenated feature is then processed through two consecutive 1×1 convolutional layers $Conv_1$ and $Conv_2$ to reduce dimensions and integrate task properties, yielding in a shared feature map denoted as $\xi _{tcl,m}$. 
\begin{equation}
\xi _{tcl,m} = \text{Conv}_1(\text{Conv}_2(\text{Concat}(\xi _{reg,m}, \xi _{seg,m})))
\label{eq:tclm}
\end{equation}

Considering the neglect of fine-grained details in the downsampling and upsampling operations, we apply the SCSE attention mechanism to improve both the shared feature map $\xi _{tcl,m}$ and the original feature maps ($\xi _{reg,m}$ and $ \xi _{seg,m}$) from the regression and segmentation branches. 
Here, the SCSE mechanism emphasizes key representations in both the spatial and channel dimensions in parallel, then combines them through addition. This process refines important features while suppressing less relevant ones.

These refined features are subsequently fused by element-wise addition, as shown by the following equations:
\begin{equation}
\xi _{tclf,m} = \text{SCSE}(\xi _{tcl,m})
\label{eq:tclfm}
\end{equation}
\begin{equation}
\xi _{regf,m} = \xi _{tclf,m} +\text{SCSE}(\xi _{reg,m})
\label{eq:regfm}
\end{equation}
\begin{equation}
\xi _{segf,m} = \xi _{tclf,m} +\text{SCSE}(\xi _{seg, m})
\label{eq:segfm}
\end{equation}
where $\xi _{regf,m}$ and $\xi _{segf,m}$ are the final obtained feature maps for each branch.

\subsection{Loss function}

During training, to jointly optimize different tasks, we design a multi-task loss function (Eq. \ref{eq:MTL}), which consists of weighted crop yield prediction loss ($L_{\text{MSE}}$), crop classification loss ($L_{\text{Dice}}$) and TCL loss ($L_{\text{TCL}}$). The overall multi-task loss function is formulated as follows:

\begin{equation}
{L_{MTL}} = a \cdot {L_{MSE}} + b \cdot {L_{Dice}} + c \cdot {L_{TCL}}
\label{eq:MTL}
\end{equation}

The crop yield decoder utilizes the Mean Squared Error (MSE) loss to minimize the difference between predicted and actual crop yields, where $ y_i$ represents the crop yield ground truth of the $i$-th point, $\hat{y_i}$ represents the predicted crop yield of the corresponding position, and $n$ can be seen as the number of labeled points in a mini-batch (Eq. \ref{eq:MSE}).

\begin{equation}
{L_{MSE}} = \frac{1}{n}\sum\limits_{i = 1}^n {{{\left( {{y_i} - {{\hat y}_i}} \right)}^2}} 
\label{eq:MSE}
\end{equation}

For crop classification, The Dice loss is employed to enhance segmentation accuracy with imbalance categories \citep{7785132}, where $X$ and $Y$ represent the ground truth and predicted map of the segmentation task. (Eq. \ref{eq:Dice}).

\begin{equation}
{L_{Dice}} = 1 - \frac{{2\left| {X \cap Y} \right|}}{{\left| X \right| + \left| Y \right|}}
\label{eq:Dice}
\end{equation}

As for the TCL task, we further consider a loss to constrain the multiscale features of the regression decoder and segmentation decoder in a unified representation space. Assume the model has a total of \textit{M} TCL blocks (in this study, \textit{M} = 5), the TCL loss $L_{TCL}$ is obtained by aggregating incorporating the TCL block of each level (Eq. \ref{eq:tcl}).

\begin{equation}
{L_{TCL}} = \sum\limits_{m = 1}^M {{{\left\| {{\xi _{regf,m}} - \xi _{tclf,m}} \right\|}^2_2} + {{\left\| {{\xi _{segf,m}} - \xi _{tclf,m}} \right\|}^2_2}}
\label{eq:tcl}
\end{equation}

Notably, the "unlabeled" part in crop yield labels and crop type labels are ignored during loss calculation.

\section{Experiments and analysis}

\subsection{Experimental settings}

\subsubsection{Benchmark methods for comparison}
To comprehensively evaluate the performance of our proposed MT-CYP-Net, we implement three classical machine learning models (Random forest \citep{Breiman2001}, XGBoost \citep{XGBoost} and LightGBM \citep{NIPS2017_6449f44a}) and two advanced deep learning models (FPN-DenseNet161 \citep{9856943} and Unet \citep{unet}), all of which are commonly used in crop yield prediction tasks. The machine learning models can predict the continuous output of crop yield based on input features such as satellite imagery reflectance value and vegetation indices. The deep learning models get single satellite imagery as input and directly output crop yield maps.

\subsubsection{Implementation details}

\textbf{Device.} All experiments were conducted on a Linux server with Pytorch an Intel i7-13600K CPU and an NVIDIA RTX 3090 GPU.

\textbf{Hyperparameters.} We use the \texttt{segmentation\_models\_pytorch} library \citep{Iakubovskii2019} to build MT-CYP-Net and Unet, employing DenseNet-161, ResNet-50 and ResNest-50d as their encoders, with the initialized ImageNet pretraining weight \citep{5206848}. All deep learning models were trained for 300 epochs using SGD with a weight decay of 0.009, a momentum of 0.9, a batch size of 8, and a cosine decay schedule with an initial learning rate of 0.008, where the warmup iteration is set to 100. For data augmentation for training, we use random horizontal flip, vertical flip, and random rotation of 90 degrees with a probability of 0.5. As for the coefficients of task weights in Eq. \ref{eq:MTL},  $a, b, c$ are configured to 5, 1, and 0.1, respectively. Related experiments can be found in the supplementary material.

\textbf{Machine learning models.} We use AutoGluon to finetune the machine learning models to ensure their best performance \citep{erickson2020autogluontabularrobustaccurateautoml}. Following previous studies \citep{DESLOIRES2023107807, QADER2023161716}, besides original pixel values, we also leverage the vegetation indices in the input, which were presented in the supplementary material.

\textbf{Deep learning models.}
To evaluate the effectiveness of the multi-task learning framework, we implement two versions of the Unet model to conduct ablation studies: one for crop yield prediction, named CY-Unet, and the other for crop classification, named CC-Unet. CY-Unet and CC-Unet share the same encoder as MT-CYP-Net, while CY-Unet using the crop yield decoder and CC-Unet using the crop classification decoder of MT-CYP-Net.

\subsubsection{Model evaluation}

We take two performance metrics, Root Mean Square Error (RMSE) and Mean Absolute Error (MAE) for crop yield prediction evaluation (Eq. \ref{eq:RMSE}, \ref{eq:MAE}), and use mean Accuracy (mAcc) and mean Intersection-over-Union (mIoU) for crop classification evaluation (Eq. \ref{eq:mIoU}, \ref{eq:mAcc}). % Notably, a lower value of RMSE and MAE indicates better crop yield prediction performance, and a higher value of mAcc and mIoU indicates better crop classification performance.

\begin{eqnarray}
\text{RMSE} = \sqrt {\frac{1}{n}\sum\limits_{i = 1}^n {{{\left( {{y_i} - {{\hat y}_i}} \right)}^2}} }
\label{eq:RMSE}
\end{eqnarray}

\begin{eqnarray}
\text{MAE} = \frac{1}{n}\sum\limits_{i = 1}^n {\left| {{y_i} - {{\hat y}_i}} \right|}
\label{eq:MAE}
\end{eqnarray}

\begin{eqnarray}
\text{mIoU} = \frac{1}{N}\sum\limits_{i = 1}^N {\frac{{TP_i}}{{TP_i + FP_i + FN_i}}}
\label{eq:mIoU}
\end{eqnarray}

\begin{eqnarray}
\text{mAcc} = \frac{1}{N}\sum\limits_{i = 1}^N {\frac{{T{P_i}}}{{T{P_i} + F{P_i}}}}
\label{eq:mAcc}
\end{eqnarray}
Where $n$ is the number of labeled crop yield points, $N$ is the number of images. $\hat{y}_i$ represents the $i$-th crop yield ground truth point, $y_i$ represents the $i$-th predicted crop yield. True Positive (TP) presents the number of positive class pixels correctly predicted. False Positive (FP) represents negative class pixels incorrectly predicted as positive. False Negative (FN) represents positive class pixels misclassified as negative.

\subsection{Comparison with benchmark methods}

\begin{table*}[h]
\caption{The performance comparison of various methods on crop yield prediction accuracy and inference efficiency using the L1C-all dataset.}
\resizebox{0.8\linewidth}{!}{%
\begin{tabular}{@{}llllll@{}}
\toprule
 & L1C-all &  & L2A-all &  & \multirow{2}{*}{Speed(s/pixel)}\\ \cmidrule(lr){2-5}
 & RMSE & MAE & RMSE & MAE &  \\ \midrule
FPN-DenseNet161\citep{9856943}& 0.2173 & 0.1337 & 0.2122 & 0.1313 &  \textbf{1.26e-09} \\
Random Forest\citep{Breiman2001} & 0.1533 & 0.0815 & 0.1575 & 0.0843 & 1.91e-07 \\
XGBoost\citep{XGBoost} & 0.1550 & 0.0761 & 0.1616 & 0.0799 & 2.12e-07 \\
LightGBM\citep{NIPS2017_6449f44a} & 0.1567 & 0.0818 & 0.1628 & 0.0843 & 3.05e-07 \\
CY-Unet(ResNet-50)\citep{unet} & 0.1578 & 0.0837 & 0.1676 & 0.0887 & \textbf{1.26e-09} \\
CY-Unet(DenseNet-161)\citep{unet} & 0.1742 & 0.0925 & 0.1759 & 0.0918 & 1.28e-09 \\
CY-Unet(ResNest-50d)\citep{unet} & 0.1543 & 0.0800 & 0.1521 & 0.0801 & 1.27e-09 \\

MT-CYP-Net(ResNet-50) & 0.1559 & 0.0821 & 0.1575 & 0.0801 & 1.28e-09 \\
MT-CYP-Net(DenseNet-161) & 0.1553 & 0.0778 & 0.1537 & 0.0784 & 1.28e-09 \\
MT-CYP-Net(ResNest-50d) & \textbf{0.1472} & \textbf{0.0706} & \textbf{0.1491} & \textbf{0.0718} & 1.28e-09 \\
\bottomrule
\end{tabular}%
}
\label{tab:Quantitative}
\end{table*}

We first quantitatively compare MT-CYP-Net with the benchmark methods and their variants in crop yield prediction. Tabel \ref{tab:Quantitative} lists the performance of different methods on both L1C-all and L2A-all datasets. MT-CYP-Net shows significantly lower RMSE and MAE values in crop yield prediction accuracy than previous methods in both L1A and L2A datasets. Moreover, we find that FPN-DenseNet161 underperforms machine learning models in our datasets, likely due to the relatively small scale of the datasets, potentially leading to overfitting issues. When using different backbone networks, MT-CYP-Net consistently demonstrates superior accuracy compared to CY-Unet. Notably, the model using ResNest-50d achieves the highest accuracy. Therefore, we choose ResNest-50d as the image encoder of MT-CYP-Net and conduct ablation experiments based on this configuration.

In addition to model performance, we also consider inference speed. Thanks to the high efficiency of the end-to-end architecture, MT-CYP-Net achieves a 149-fold improvement over the fastest machine learning model, while being only 1.6\% slower than CY-Unet and FPN-DenseNet161. These results demonstrate that MT-CYP-Net achieves an excellent balance between precision and efficiency.

\begin{figure*}[h]
    \centering
    \includegraphics[width=1\linewidth]{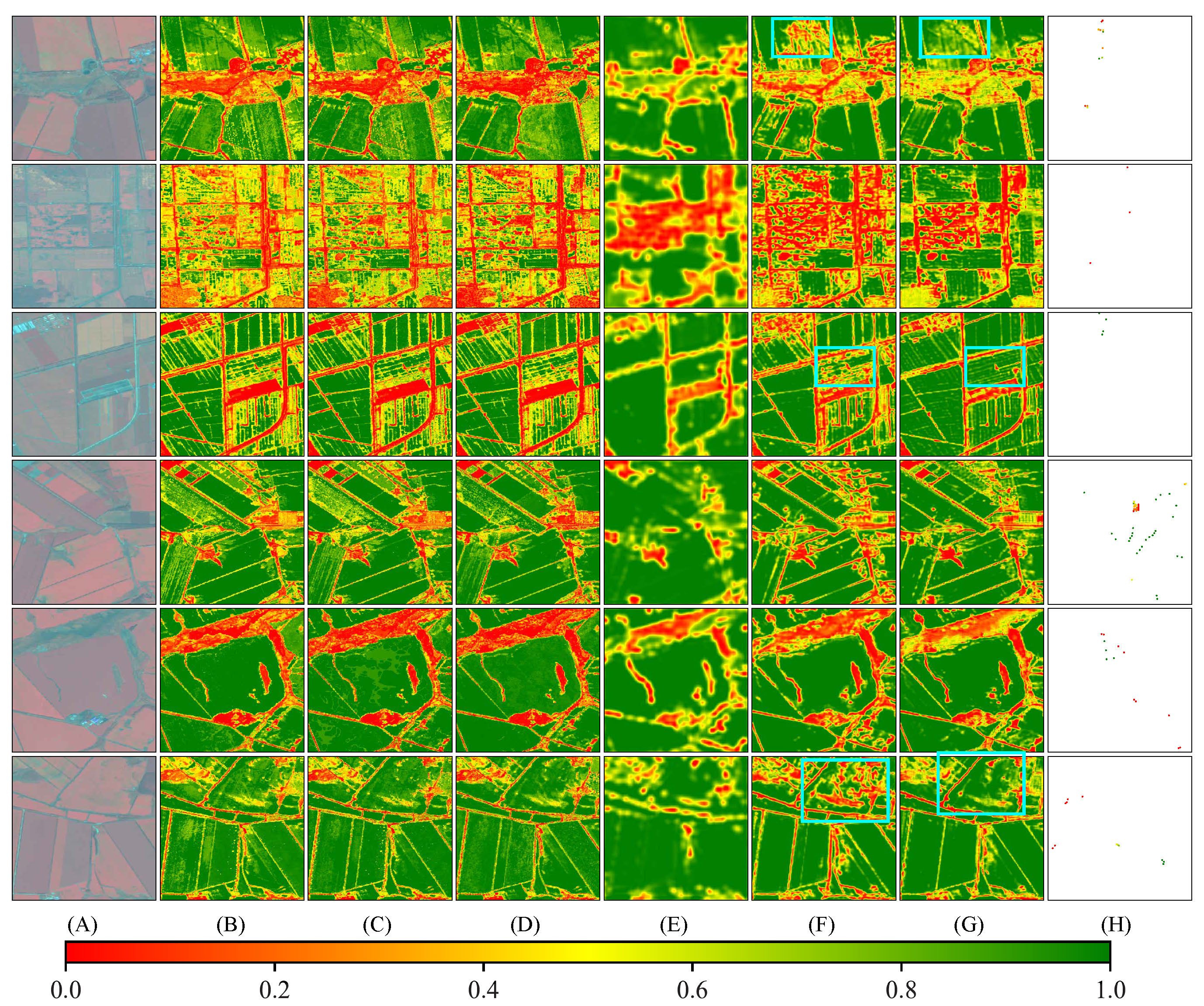}
    \caption{Visualization of crop yield prediction results in L1C-all dataset. (A) Sentinel-2 L1C images displayed in pseudo color (NIR, Green, Blue); (B) Random Forest; (C) XGBoost; (D) LightGBM; (E) FPN-DenseNet161; (F) CY-Unet; (G) MT-CYP-Net; (H) Crop yield point label.}
    \label{fig:Visialization-compare}
\end{figure*}

Fig. \ref{fig:Visialization-compare} displays the crop yield prediction maps of different methods. As can be seen, the crop yield prediction results of the three machine learning models exhibit consistency. Their outputs are sharper and more granular compared to CNN models. However, since these methods do not consider contextual information, they generate numerous pixel-level noise.

Compared to machine learning methods, the CNN models show excessively smooth predictions. However, aligning to the statistical indices of Table \ref{tab:Quantitative},  FPN-DenseNet161 performs poorly in qualitative visualizations. We speculate that this is due to overfitting issues arising from the use of a heavy backbone network trained on a limited dataset. In contrast, CY-Unet and MT-CYP-Net have a stronger recognition ability in most cases, where MT-CYP-Net performs better than CY-Unet at the edge of the field. In addition, without the assistance of category information, CY-Unet often shows an overestimation of the extent of damaged areas (see the cyan box in Fig. \ref{fig:Visialization-compare}), highlighting the significance of multi-task collaboration training.

\subsection{Ablation studies}\label{sec:Ablation}

\begin{table*}[h]
\caption{Ablation experiment results of MT-CYP-Net on different tasks.}
\centering
\resizebox{0.8\linewidth}{!}{%
\begin{tabular}{@{}llllll@{}}
\toprule
 & Crop yield prediction &  &  & Crop classification &  \\ \cmidrule(lr){2-3} \cmidrule(l){5-6} 
 & RMSE & MAE &  & mIoU & mAcc \\ \midrule
CY-Unet & 0.1543 & 0.0800 & CC-Unet & 79.6241 & 87.5156 \\
MT-CYP-Net-hard & 0.1513 & 0.0730 & MT-CYP-Net-hard & 81.2297 & 88.3870 \\
MT-CYP-Net & \textbf{0.1472} & \textbf{0.0706} & MT-CYP-Net & \textbf{81.6942} & \textbf{88.7097} \\ \bottomrule
\end{tabular}%
}
\label{tab:mt-ablation}
\end{table*}

\subsubsection{Multi-task structure and TCL block}
Next, we conduct an ablation study to demonstrate the necessity and effectiveness of the multi-task structure and TCL block of MT-CYP-Net. Hard parameter sharing and soft parameter sharing are two widely used multi-task learning methods in dense prediction  \citep{9336293}. MT-CYP-Net implements soft parameter sharing by integrating the TCL block. To assess the impact of each component on the model performance, we first remove the TCL blocks from MT-CYP-Net, creating a hard parameter sharing variant (MT-CYP-Net-hard). Subsequently, we conduct ablation studies on vanilla Unet (including CY-Unet and CC-Unet), MT-CYP-Net-hard, and MT-CYP-Net. The results in Table \ref{tab:mt-ablation} show that MT-CYP-Net-hard outperforms vanilla Unet in both crop yield prediction and crop classification tasks, showing the advantage of MTL. Moreover, MT-CYP-Net achieves better performance compared to MT-CYP-Net-hard, demonstrating that the TCL module is able to promote mutual complementarity between tasks, aligning the features of crop type classification and crop yield prediction into a unified space.

\begin{figure*}[t]
    \centering
    \includegraphics[width=1\linewidth]{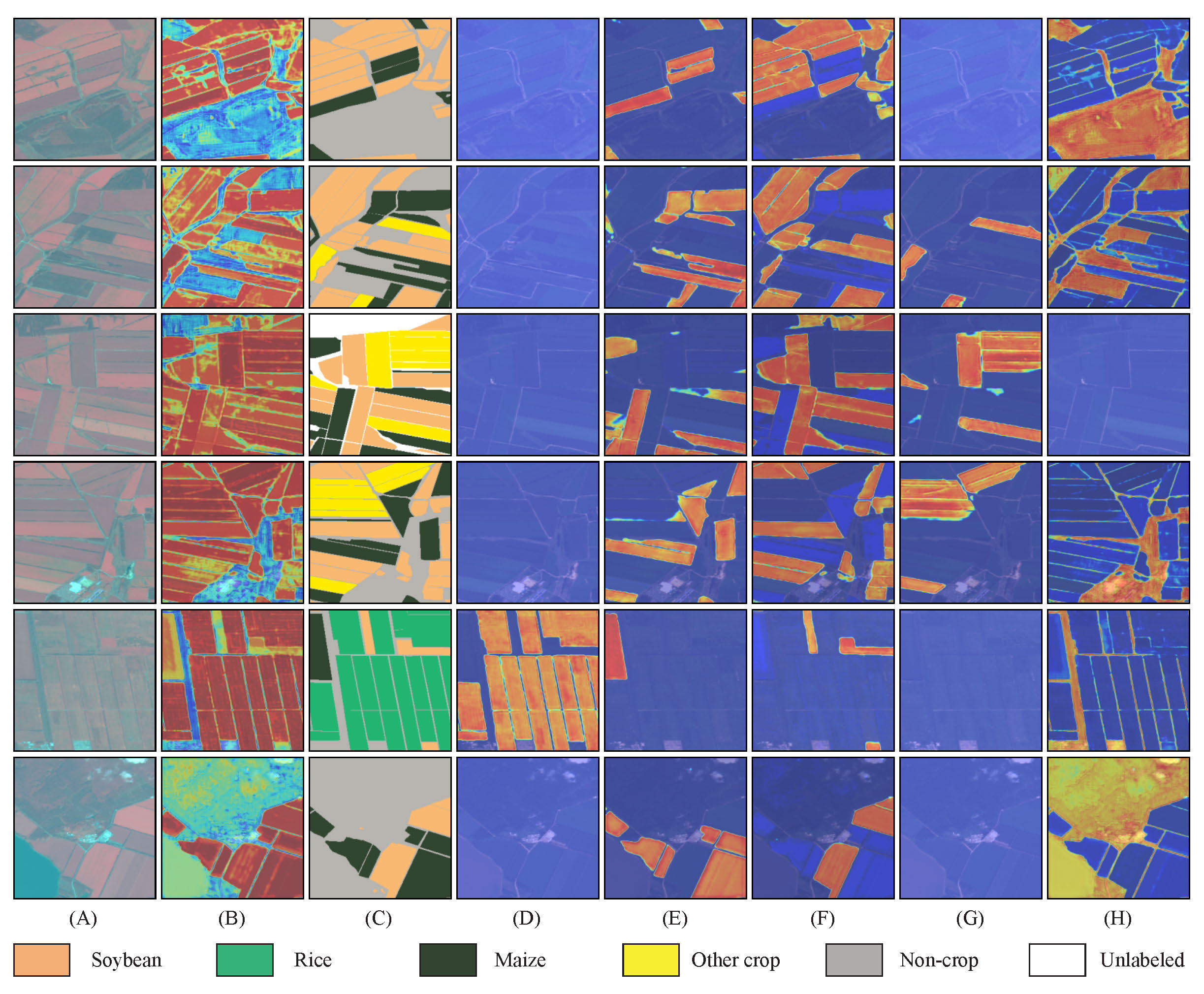}
    \caption{Grad-CAM visualization results of the refined shared feature map of the final TCL block in MT-CYP-Net. (A) Sentinel-2 L1C images displayed in pseudo color (NIR, Green, Blue); (B) CAM on crop yield; (C) Crop type label; (D) CAM on rice; (E) CAM on maize; (F) CAM on soybean; (G) CAM on other crop; (H) CAM on non-crop.}
    \label{fig:cam}
\end{figure*}

To further analyze these results, we apply Grad-CAM \citep{8237336} to the refined shared feature map of the final TCL block ($\xi _{tclf,5}$, see in Fig. \ref{fig:network}) and visualize the related responses of the predicted crop yield from regression decoder and each category generated from segmentation decoder, as shown in Fig. \ref{fig:cam}. The CAM of crop yield closely aligns with the intensity of red in the pseudo-color images, which reflects the vegetation growth status (Fig. \ref{fig:cam} AB). For the crop classification task, the attention areas corresponding to rice, maize, soybeans, and other crop and non-crop regions are distinctly presented (Fig. \ref{fig:cam} DEFGH). In addition, we can also observe that the red regions of the CAM in the crop yield prediction task are the union of the highlighted regions of different crops. These findings demonstrate that the TCL block can effectively capture and utilize the shared information between tasks.

\subsection{How does MT-CYP-Net perform across different crop types?}

\begin{figure}[t]
    \centering
    \includegraphics[width=0.5\linewidth]{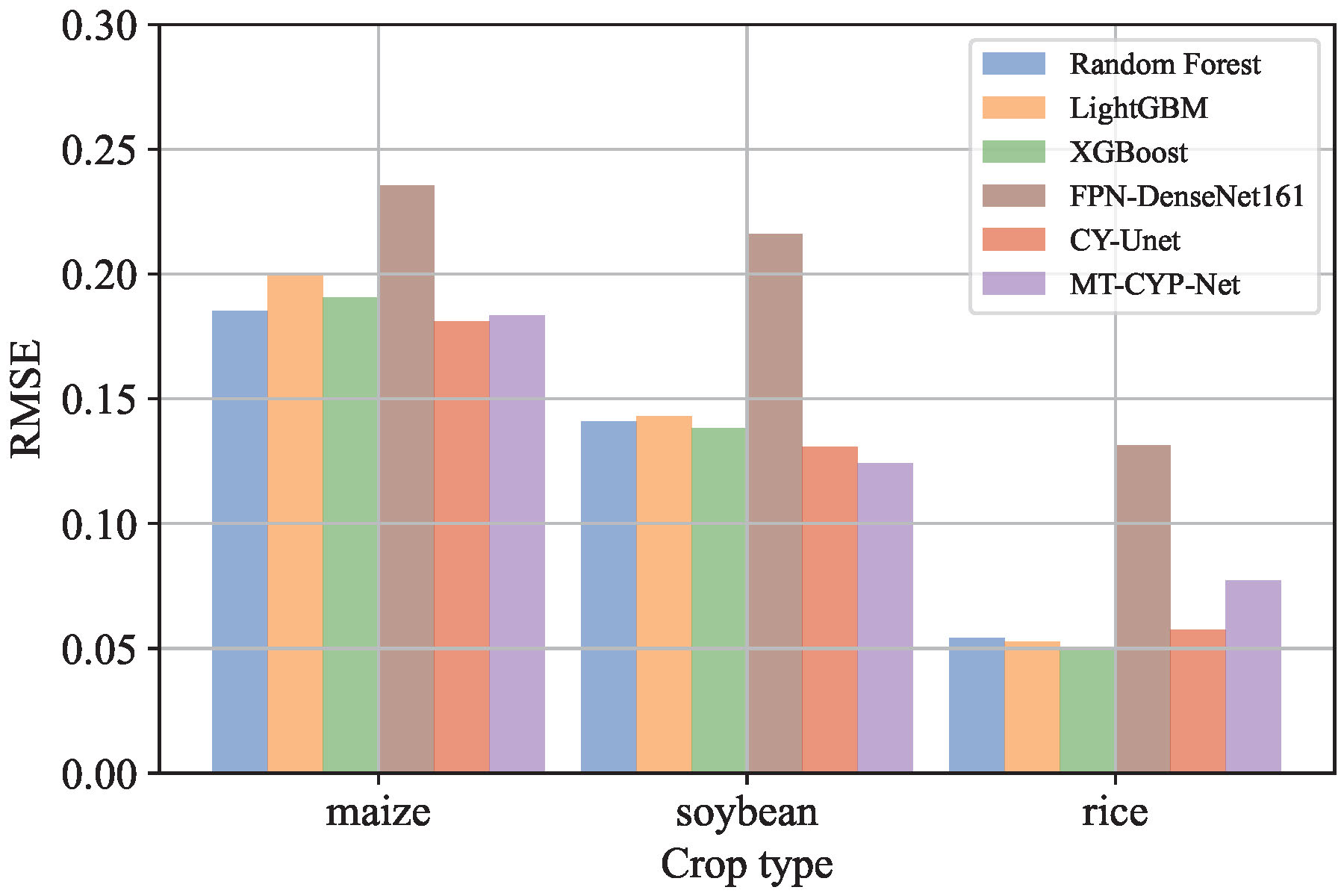}
    \caption{The crop yield prediction accuracy of different crop types.}
    \label{fig:RMSE-crop}
\end{figure}

In this section, we further analyze the model's performance across different crop types, using L1C-all bands for convenience. From Fig. \ref{fig:RMSE-crop}, we observe a consistent trend between machine learning and CNN models: rice has the lowest error, followed by soybean, with maize exhibiting the highest error. We attribute this difference to the smaller yield variations of rice compared to other crop types in the context of flood disasters (see Section \ref{sec:discuss} for more details).

\subsection{How does MT-CYP-Net perform with different band combination input?}

\begin{figure}[h]
    \centering
    \includegraphics[width=0.5\linewidth]{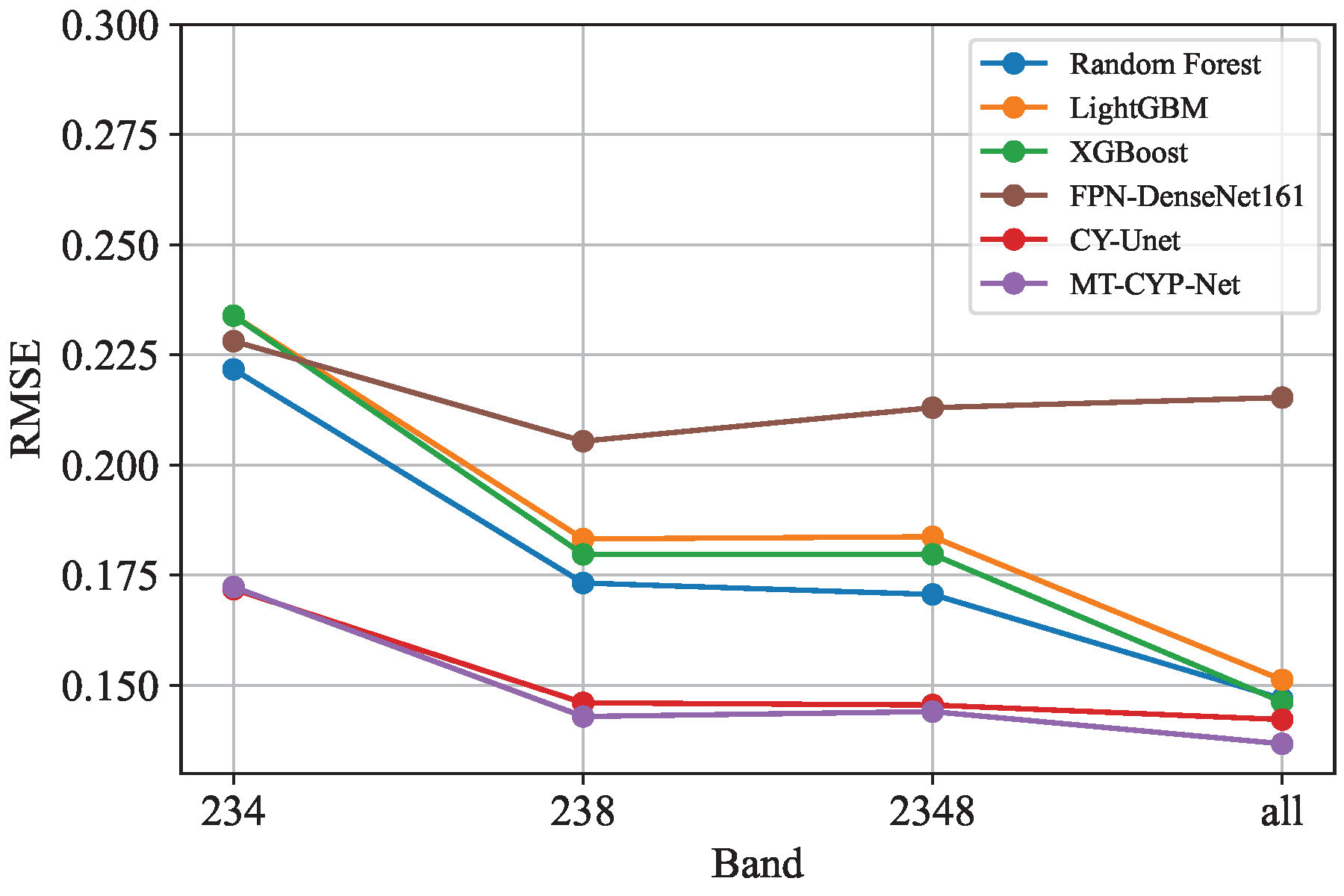}
    \caption{The RMSE value of crop yield prediction results in different spectral band combination inputs.}
    \label{fig:L1C-RMSE}
\end{figure}

To further evaluate the generalization capability of the models, we test their performance using different combinations of spectral bands. Here, since we mainly consider the channel with a resolution of 10m %(Table \ref{tab:Sentinel-2-band})
, besides directly adopting all bands (L1C-all and L2A-all), only the bands of B02, B03, B04 and B08 are employed, obtaining the following combinations: band 2,3,4, band 2,3,8 and band 2,3,4,8. Fig. \ref{fig:L1C-RMSE} shows MT-CYP-Net consistently outperforms other models across different spectral band combinations. In general, the accuracy of crop yield predictions improves as the number of input bands increases, with all models achieving their highest performance when utilizing all available bands, indicating the effectiveness of spectral information for understanding the growth status of crops.

Notably, the accuracy of MT-CYP-Net is relatively less sensitive to the number of input bands compared to traditional machine learning models. It is possibly because compared to spectral information, spatial information is more important for CNNs. Therefore, even in scenarios with only three input bands  (e.g., 234, 238), MT-CYP-Net still can significantly outperform machine learning models, while the gaps are gradually narrowed with the utilization of more channels.

\subsection{How does MT-CYP-Net perform in few-shot learning scenarios?}

\begin{figure}[h]
    \centering
    \includegraphics[width=0.6\linewidth]{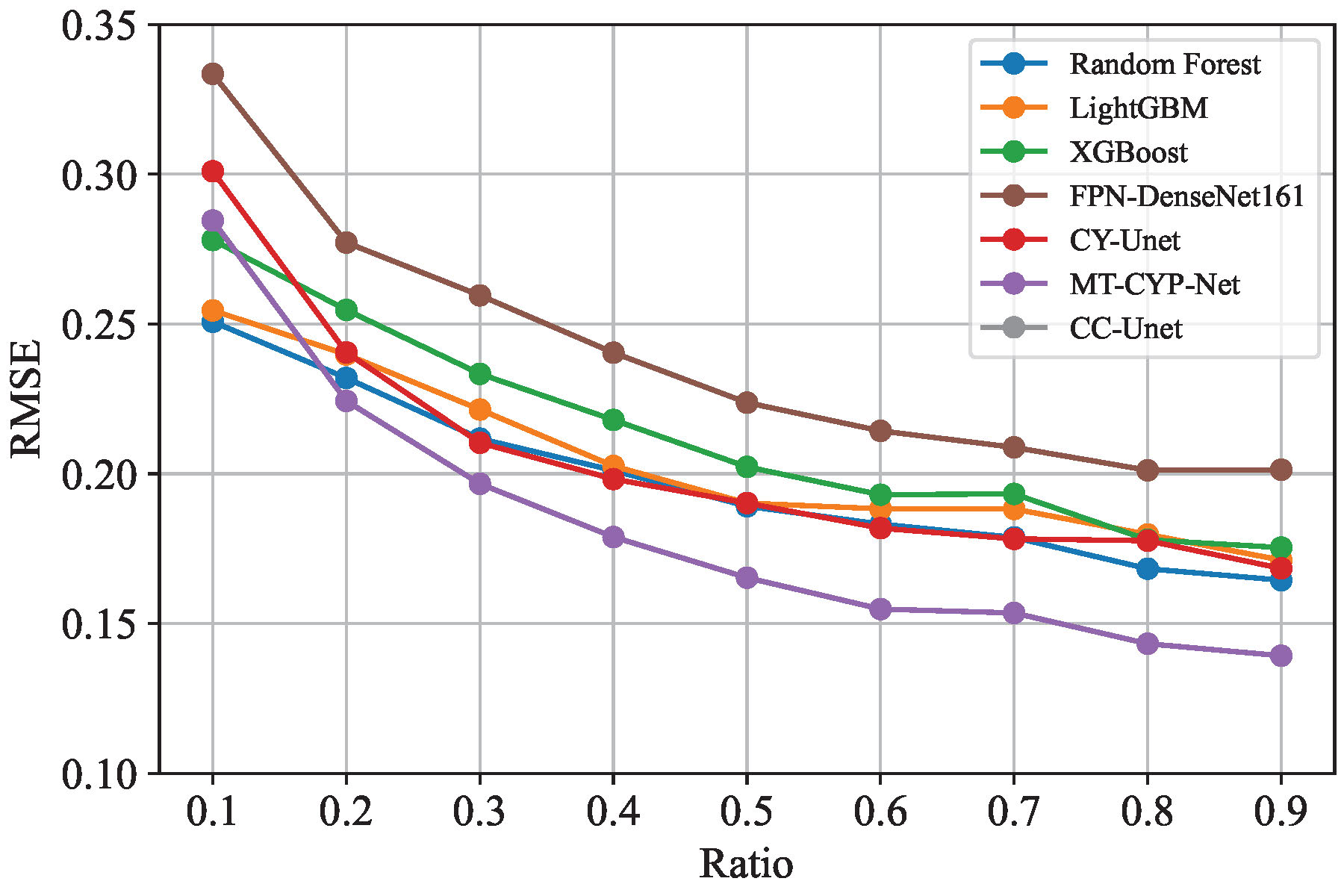}
    \caption{The few-shot performance of models. The RMSE value is the average of the results from ten experiments.}
    \label{fig:fewshot}
\end{figure}

Collecting crop yield data is usually expensive and challenging, making few-shot performance critical for evaluating crop yield prediction models. Therefore, we further evaluate the few-shot performance of the six models using the L1C dataset. In this experiment, it is divided into training and validation sets with a 7:3 ratio. We train and validate the models using varying portions of the training set, ranging from 10\% to 90\%. This process was repeated 10 times with different random seeds to ensure robustness. As illustrated in Fig. \ref{fig:fewshot}, for all models, the RMSE value consistently decreases as the training ratio increases from 0.1 to 0.9, indicating improved model performance with more training data. It is worth noting that the machine learning models outperform CNN-based models when only 10\% of the training data is used, suggesting that machine learning models are able to learn a discriminative space only with few data, while the deep networks are still underfitting at this time. Nevertheless, as the training ratio increases, CNN models show greater performance improvements, eventually surpassing machine learning models. Notably, our MT-CYP-Net consistently outperforms CY-Unet and FPN-DenseNet161 across all sample ratio scenarios and maintains the highest accuracy once the training ratio exceeds 20\%.

\begin{comment}

\subsection{How does MT-CTP-Net perform on the crop classification task?}

\begin{figure}[tbp]
    \centering
    \includegraphics[width=\linewidth]{fig/classification_compare.jpg}
    \caption{Visualization of crop classification results in L1C-all dataset. (A) Sentinel-2 L1C images displayed in pseudo color (NIR, Green, Blue); (B) CC-Unet; (C) MT-CYP-Net; (D) Crop type labels.}
    \label{fig:classification_compare}
\end{figure}

Although crop classification in MT-CYP-Net is the auxiliary task for crop yield prediction, it still achieves higher accuracy than CC-Unet (Table \ref{tab:mt-ablation}). In view of this, we further qualitatively compare the crop classification results of MT-CYP-Net and CC-Unet in Fig. \ref{fig:classification_compare}. It can be seen that, MT-CYP-Net has a better visualization than CC-Unet. For example, CC-Unet often misclassifies field ridges and roads as crop categories, while MT-CYP-Net performs well in this regard (see the red boxes). In addition, from Fig. \ref{fig:Visialization-compare} and Fig. \ref{fig:classification_compare} we can find that, MT-CYP-Net possesses notable edge consistency in the crop yield prediction and classification results, highlighting the advantages of mutual improvement between tasks in MTL.

\end{comment}

\section{Discussion}\label{sec:discuss}

\subsection{Application visualization in farm-scale}

\begin{figure*}[htbp]
    \centering
    \includegraphics[width=1\linewidth]{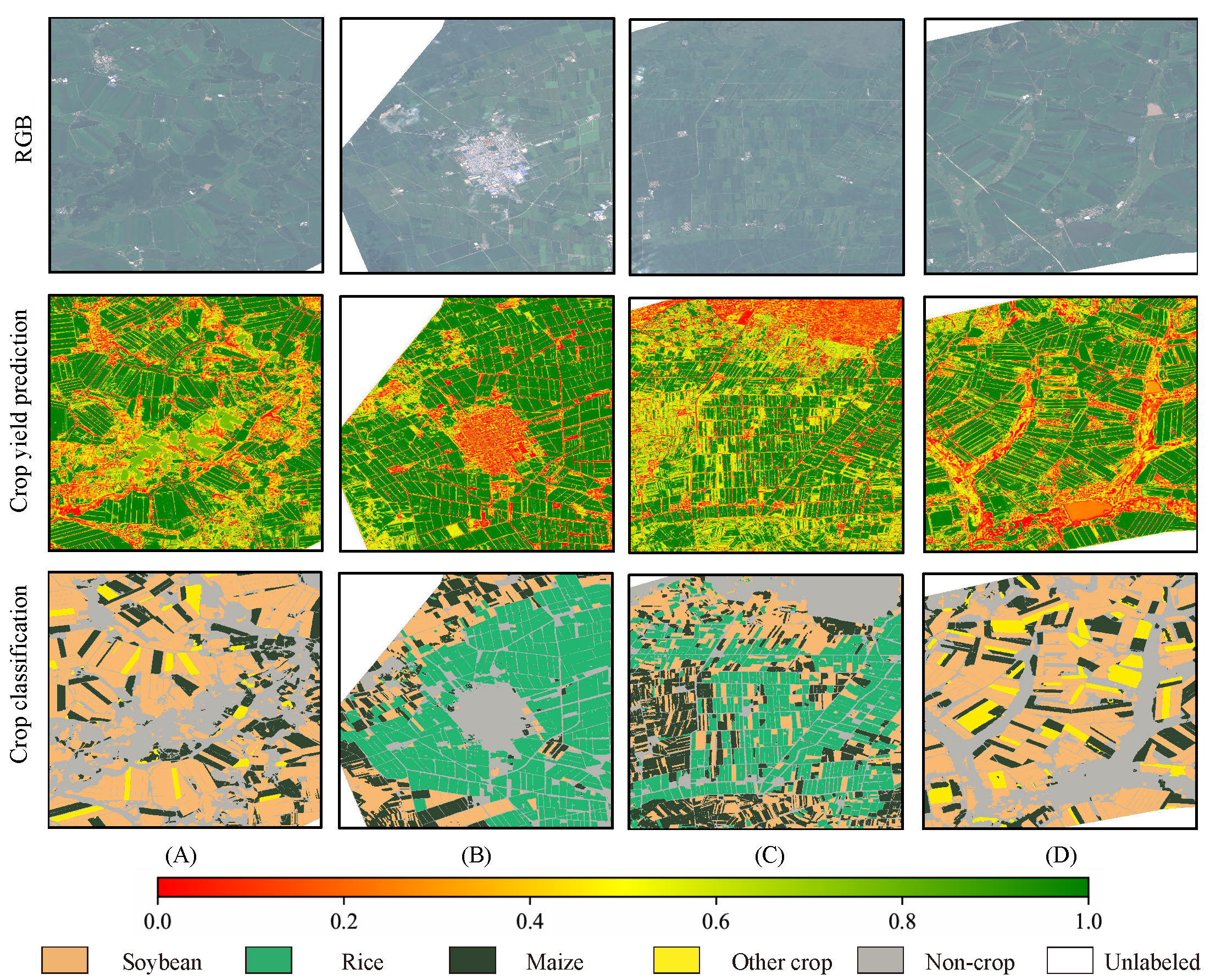}
    \caption{The farm-scale crop yield prediction and classification results. (A) Rongjun; (B) Qixing; (C) Wujiuqi; (D) Heshan.}
    \label{fig:farm-scale}
\end{figure*}

In this section, we expand MT-CYP-Net to demonstrate its feasibility and generalizability  for larger-scale farm-level mapping. We selected four farms located in different regions of Heilongjiang Province: Rongjun, Qixing, Wujiuqi, and Heshan, to showcase their farm-scale crop yield prediction and crop classification mapping. These farms represent diverse geographic and agronomic conditions. Rongjun and Heshan are situated in the northern part of the Songnen Plain, with soybeans and maize as the primary crops. In contrast, Qixing and Wujiuqi are located in the central Sanjiang Plain, where rice, soybeans, and maize are the main crops.

Fig. \ref{fig:farm-scale} illustrates the crop yield and type distributions across large areas using pixel-level high-resolution maps. The classification results exhibit strong plot integrity, effectively identifying non-crop areas such as rivers, reservoirs, and urban regions, where yields are near zero. Nevertheless, this method may erroneously indicate relatively high yields in forested regions. Overall, the farm-scale visualizations demonstrate that MT-CYP-Net can achieve efficient and large-scale crop yield prediction with low data collection costs, highlighting the practicality and generalizability of the proposed solution.

\subsection{Limitation and prospections}
This study proposes a deep-learning method for crop yield prediction using a CNN model under very few crop yield samples, to realize crop yield prediction based on satellite images. Although it achieves the best performance in our dataset compared to %classical machine learning methods and previously used CNN models
its counterparts, it still has some limitations to improve in the future.  

\textbf{Single temporal.} The heterogeneity of crop growth across species makes it challenging to accurately predict crop yield based solely on single temporal images. In our study area, each crop type includes dozens of varieties with distinct canopy structures and growth characteristics. This heterogeneity may introduce bias in the interpretation based on a single temporal remote sensing image model. Satellite Image Time Series (SITS) fusion can capture the temporal dynamics of variety-specific crop growth characteristics, which we hope to explore in the future. 

\textbf{Data imbalance.}  Limited by manpower and time, the crop yield distribution in our dataset is highly skewed, with most samples concentrated in high-yield regions close to 1 and low-yield regions close to 0, while middle-range yield data is scarce. Consequently, this study does not evaluate the models' performance across different yield ranges, as such analysis would be more appropriate for datasets with a uniform distribution \citep{Ren_2022_CVPR, pmlr-v139-yang21m, 10.1007/978-3-031-20044-1_4}. Nevertheless, it is necessary to acknowledge the discrepancy between the actual crop yield distribution and the long-tailed distribution of our dataset. The imbalance presents potential biases that should be considered when interpreting the results. 

\textbf{Single modality.} Meteorological conditions affect the availability of satellite data, thereby restricting the application of the model. The rainy weather usually poses challenges for acquiring optical satellite imagery. In such conditions, Synthetic Aperture Radar (SAR) can be considered. Although previous studies have applied SAR images to crop yield prediction, its multi-modal fusion with optical satellite data needs further exploration.

\textbf{Natural disaster.} Our study area frequently experiences flooding and waterlogging events, often persisting until September. The satellite images are generally captured on clear days following disaster events. At this time, the oxygen stress experienced by plants may not yet be reflected in the canopy's spectral characteristics \citep{YANG20212613}. Therefore, the response of crop canopies to waterlogging has a delayed effect, with the consequences becoming visible several weeks after the event. This temporal lag creates a discrepancy between reflectance data captured by satellites and actual crop fields.

\section{Conclusion}

% Accurate large-scale crop yield prediction is essential for ensuring global food security. However, current technologies fall short of achieving this goal, posing significant risks to agricultural sustainability. To solve this problem, 
In this study, we propose MT-CYP-Net, a CNN-based model designed for predicting crop yield with minimal data collection costs. MT-CYP-Net leverages crop yield prediction and crop classification tasks in a unified end-to-end MTL framework, where the features extracted from a backbone network are utilized by different task-specific decoders, and the multi-scale feature maps in the multi-task decoder interact and fuse through well-designed TCL blocks. This design enables the model to learn better features, significantly reducing the data annotation cost while enhancing overall performance. To support this approach, %we constructed a dataset using satellite images from eight farms in Heilongjiang Province, China. The dataset includes crop yield-labeled maps and category distribution maps, with crop yield labeled at the point level on a very limited number of pixels to minimize annotation efforts.
we created a dataset from satellite images of eight farms in Heilongjiang Province, China, including minimal point-level yield and crop category maps to reduce annotation efforts. Extensive quantitative comparisons demonstrate that MT-CYP-Net achieves competitive performance in both accuracy and inference speed compared to classical machine learning models and existing deep learning methods. The ablation studies also demonstrate the necessity and effectiveness of the multi-task structure and TCL block in MT-CYP-Net. When expanding to farm-level applications, it shows a strong generalization ability and demonstrates significant application value on large-scale geographic region mapping, underscoring its potential to enhance agricultural management and food security planning.

% \section*{Acknowledgments}
% This work was funded by the Program of the National Natural Science Foundation of China (NSFC) (grant numbers 52379045, and 52179039).

% \appendix

% \section{Vegetation indices used in machine learning models}
% \label{sec:sample:appendix}

\clearpage
%\bibliographystyle{elsarticle-harv} 
\bibliographystyle{cas-model2-names}
\bibliography{cas-refs}





% \end{thebibliography}
\end{document}